\documentclass[a4paper, 10 pt, conference]{ieeeconf} 

\IEEEoverridecommandlockouts  
\usepackage{graphicx}
\usepackage{amsmath,amssymb} 
\usepackage{color}
\usepackage{subcaption}
\usepackage[linesnumbered,ruled,vlined]{algorithm2e} 
\usepackage{multicol}
\usepackage{bm}
\usepackage[hyphens]{url}
\usepackage{multirow}
\usepackage{times}
\usepackage{gensymb}
\usepackage{cite}
\usepackage{array}
\usepackage[bookmarks=false]{hyperref}
\usepackage[font={small,it}]{caption}
\hypersetup{
    colorlinks=true,     
    urlcolor=cyan,
}

\definecolor{rColor}{rgb}{0, 0, 1}

\definecolor{bColor}{rgb}{1, 0, 0}

\definecolor{gColor}{rgb}{0, 1, 0}

\graphicspath{{./figures/rails_ims/}{./figures/valve_ims/}{./figures/ac1_ims/}{./figures/}{./figures/classification/our_errors/}{./figures/classification/resnet_errors/}} 

\title{\LARGE \bf FaultNet: Faulty Rail-Valves Detection using Deep Learning and Computer Vision}
\author{Ramanpreet~Singh~Pahwa$^{1}$,
		Jin~Chao$^{1}$,
		Jestine~Paul$^{1,2}$, 
		Yiqun~Li$^{1}$,
		Ma~Tin~Lay~Nwe$^{1}$, 
		Shudong~Xie$^{1}$, \\
		Ashish~James$^{1}$,
		Arulmurugan~Ambikapathi$^{1}$, 
		Zeng~Zeng$^{1}$, and 
		Vijay~Ramaseshan~Chandrasekhar$^{1}$%
\thanks{$^{1}$R.~S.~Pahwa, J.~Chao, J.~Paul, Y.~Li, T.~L.~N.~Ma, S.~Xie, A.~James, A.~Ambikapathi, Z.~Zeng, and V.~R.~Chandrasekhar are with Institute for Infocomm Research (I$^2$R), A*STAR, Singapore. }%
\thanks{$^{2}$J.~Paul is with National University of Singapore (NUS), Singapore. }%
}

\begin{document}
\maketitle
\thispagestyle{empty}
\pagestyle{empty}
\begin{abstract}
Regular inspection of rail valves and engines is an important task to ensure safety and efficiency of railway networks around the globe. Over the past decade, computer vision and pattern recognition based techniques have gained traction for such inspection and defect detection tasks. An automated end-to-end trained system can potentially provide a low-cost, high throughput, and cheap alternative to manual visual inspection of these components. However, such systems require huge amount of defective images for networks to understand complex defects. In this paper, a multi-phase deep learning based technique is proposed to perform accurate fault detection of rail-valves. Our approach uses a two-step method to perform high precision image segmentation of rail-valves resulting in pixel-wise accurate segmentation. Thereafter, a computer vision technique is used to identify faulty valves. We demonstrate that the proposed approach results in improved detection performance when compared to current state-of-the-art techniques used in fault detection.
\end{abstract}

\begin{section}{Introduction}
Railway systems are among the most used and preferred public transportation system around the globe. Singapore MRT, on an average, transports three million passengers per day. This has been more than a $200\%$ increase in ridership in the last 15 years \cite{mrt_website}. The number of trains, speed, and operating hours of trains have also increased significantly over the years. All these factors inevitably raise the risk of trains breaking down which can lead to endangering human life and significant loss in countries' productivity and economy. There are multiple reasons for rail surface defects such as fatigue, impacts from damaged wheels, and repetitive passing of rolling stock over rail components such as welds, joints, and switches. If these defects are not detected in time, it can lead to accidents resulting in significant loss of citizens' lives and revenue.

Machine Learning (ML) and Deep Learning (DL) are fast becoming an integral part of advanced manufacturing and inspection fields such as component design \cite{wang2018deep}, optical inspection \cite{weimer2016design}, predictive maintenance \cite{li2017intelligent}, and anomaly detection \cite{zenati2018efficient}. Such systems, with the availability of vast datasets, have improved over classical handcrafted feature learning approaches significantly in various domains. Some of the improvements made in these domains are also applicable to fault inspection and detection in railway networks.

Non-destructive testing (NDT) techniques are used for some defect detection and predictive maintenance. Some of the available NDT techniques for railway inspection use color cameras, eddy currents, and ultrasonics \cite{Papaelias2008, CLARK2004111}. Ultrasonics has results in one of the best performance for detecting internal rail cracks \cite{Clark2002341, EDWARDS2006468}. However, its inspection speed is slow and is unable to detect surface defects. Several techniques such as electromagnetic acoustic transducers, lasers, and air-coupled ultrasonics have been proposed to improve the detection speed, but these attempts have largely been unsuccessful \cite{Papaelias2008}. Another NDT technique, eddy current, uses magnetic field generated by eddy currents for defect identification \cite{Bentoumi2003}. Eddy current has relatively high inspection speed and thus, it is often combined with ultrasonics for rail inspection. However, its sensor is very sensitive to the lift-off variation with the probe positioned at a constant distance (no more than $2$mm) from the surface of the rail head \cite{Thomas2007}. 

\begin{figure*}[t!]
\centering
\includegraphics[width=0.95\textwidth]{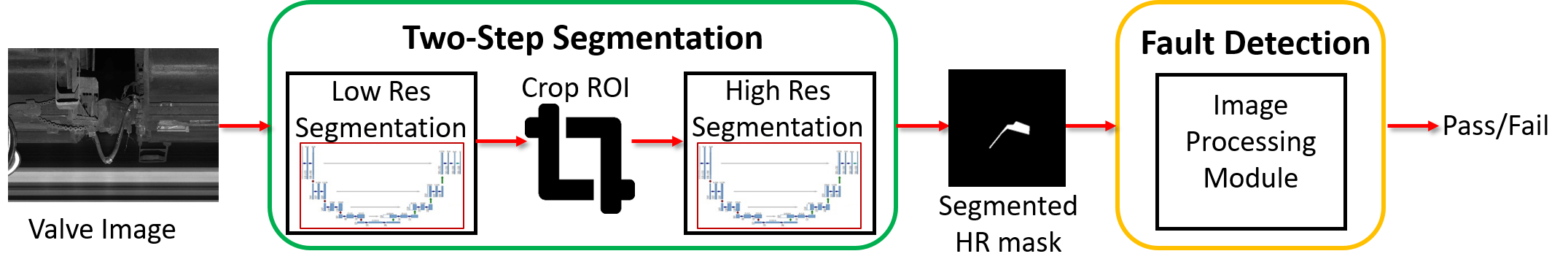}
\caption{FaultNet architecture. We perform a two-step high resolution segmentation of the train valves and use image processing techniques to identify faulty valves.}
\label{fig::overview}
\end{figure*}
Recently, various Visual Track Inspection System (VTIS) for rail surface detection has been developed utilizing the tremendous advances in computer vision techniques. In VTIS, a high speed camera is installed under a train or on the side walls of the train tunnels. It captures the images of the train tracks and both train-sides. Further analysis of the captured images is usually performed by an image processing software for custom applications such as bolt detection \cite{Marino2007}, corrugation inspection \cite{Mandriota2004}, and crack detection \cite{LinJie2009}. Such visual inspection systems have the potential to be fast and low cost making them very attractive for track surface defect detection \cite{Papaelias2008}. However, currently many of the commercial-off-the-shelf (COTS) VTIS systems have high false alarm rates or are incapable of detecting subtle defects such as faulty valves. This results in a significant amount of maintenance man-hours spent to visually analyze every image to screen for these types of defects.

This paper focuses on the VTIS system and proposes a multi-phase deep learning based valve defect detection technique as shown in Fig.~\ref{fig::overview}. We make two main contributions in the proposed hybrid architecture:
\begin{itemize} 
\item We perform high resolution image segmentation using a two-step DL-based segmentation approach in the first phase. 
\item We develop a custom computer vision technique to process the segmented mask in first phase to detect faulty valves.
\end{itemize}

Unlike conventional segmentation techniques, our two-step segmentation approach ensures that we obtain accurate segmentation of the rail valves from the raw images. It also benefits the classification being performed in the final phase by discarding the noisy background and other non-relevant information. The cropped segmentation of the Region of Interest (ROI) is then used for the final fault detection phase where we use custom-built computer vision techniques to accurately identify faulty rail-valves.

Our paper is structured as follows. In Section \ref{sec::related_works}, we review other works that are related to our research topic. In Section \ref{sec::overview}, we provide a brief overview of our problem formulation. Section \ref{sec::our_approach} presents the proposed segmentation and fault detection steps in detail, highlighting our contributions and observations at each stage. We report our results and comparison with other state-of-the-art DL techniques in Section \ref{sec::results}. Finally, in Section \ref{sec::conclusion} we conclude this paper and discuss future research directions.

\end{section}
  
\begin{section}{Related Work}\label{sec::related_works}

One of the most important tasks in Computer Vision field is of automated image segmentation, specifically background and foreground segmentation. Various approaches \cite{shi2000normalized, felzenszwalb2004efficient, haralick1985image, arbelaez2011contour, grady2006random} have been used to perform this task quickly and more accurately. Usually, an input image is analyzed using different features such as color, size, and histograms at various scales and a foreground-background (denoted in white and black respectively) region is obtained. With the advent of deep learning, various techniques such as U-Net \cite{ronneberger2015u}, SegNet \cite{badrinarayanan2015segnet}, mask-RCNN \cite{maskrcnn} have improved over the traditional hand-crafted techniques. U-Net based segmentation has quickly become the industry standard for binary segmentation related tasks. It consists of a contracting encoder to analyze the entire image followed by an expanding decoder to produce an accurate segmentation of the object of interest. This property has enabled U-Net based architectures to win various benchmark competitions on bio-medical segmentation and detection competitions. The standard U-Net architecture usually downsamples the entire image to a smaller resolution such as $256\times256$ and performs image segmentation on the downsampled data. Thereafter, the segmented masks are upsampled to the original resolution to output the final segmentation. The initial downsampling of the input image results in loss of important information, especially at edges, that is critical for accurate segmentation. In this work, we address this issue by performing a two-step high resolution segmentation that first computes an ROI and then crops this region for performing segmentation on full (and cropped) resolution for a more accurate segmentation of the object of interest.

Vision based track inspection technology has been gradually adopted by the railway industry since the pioneering work by Torsino et. al. \cite{Torsinopatent1, Torsinopatent2}. Classically, the common choices of features for detection from visual data have been histogram of oriented gradients (HoG), Scale-Invariant Feature Transforms (SIFT), spacial pyramids, and Gabor filters. A two 3-layer neural network running in parallel is used to detect hexagonal headed bolts in \cite{Marino2007, Ruvo20092333}. Other approaches include using Support Vector Machine (SVM) classifier \cite{Gilber2007} and an AdaBoost-based object detector \cite{YLi2014} to identify and classify cracks, tie plates, and missing spikes. A Convolutional Neural Network (CNN) followed by ResNet is trained on a database of train tracks for detecting defects on track surfaces in \cite{ashish_tracknet}. In \cite{Gibert2017, Giben2015, RoohiIJCNN}, varying deep CNN architectures have been used for monitoring, identification, and classification of rail surface defects and railway-track components. Such techniques work well when we have thousands of groundtruth images to train the deep classification networks. Data augmentation is generally used to augment limited or sparse datasets. It is critical that our classification results do not have any skips (false-positives). This is to ensure that no faulty valves are incorrectly detected to be normal valves. Our experiments using the current-state-of-the-art classification techniques resulted in an unacceptable amount of skips. Thus, we develop a custom computer-vision based technique to identify faulty-valves that ensures as minimal skips as possible. We demonstrate our technique's superior performance over ResNets and DenseNets in the results section.

\end{section}

\begin{section}{Problem Description}\label{sec::overview}
The images are captured by an image acquisition system and fed into an image analysis system for various rail surface defect detection and classification. For certain defects, where current COTS aren't available, a human reviewer looks at the images and uses a computer-vision aided software to make an executive decision for such defects. This makes the task extremely cumbersome and prone to human errors having to inspect huge amount of images every day. In this work, we automate the process of fault detection for rail-valves. The valve's main body needs to be angled downwards towards the long valve-handle as shown in Fig.~\ref{fig::valve_example_crop}(a). If the main body is horizontal or angling above the long valve-handle, it is considered a fault valve as shown in Fig.~\ref{fig::valve_example_crop}(b). Our proposed deep learning technique, FaultNet, automates the fault detection process and improves the performance over current state-of-the-art techniques as described in the following section.

\begin{figure}[tpb]
\centering
\begin{subfigure}[b]{0.225\textwidth}
\includegraphics[width=\textwidth]{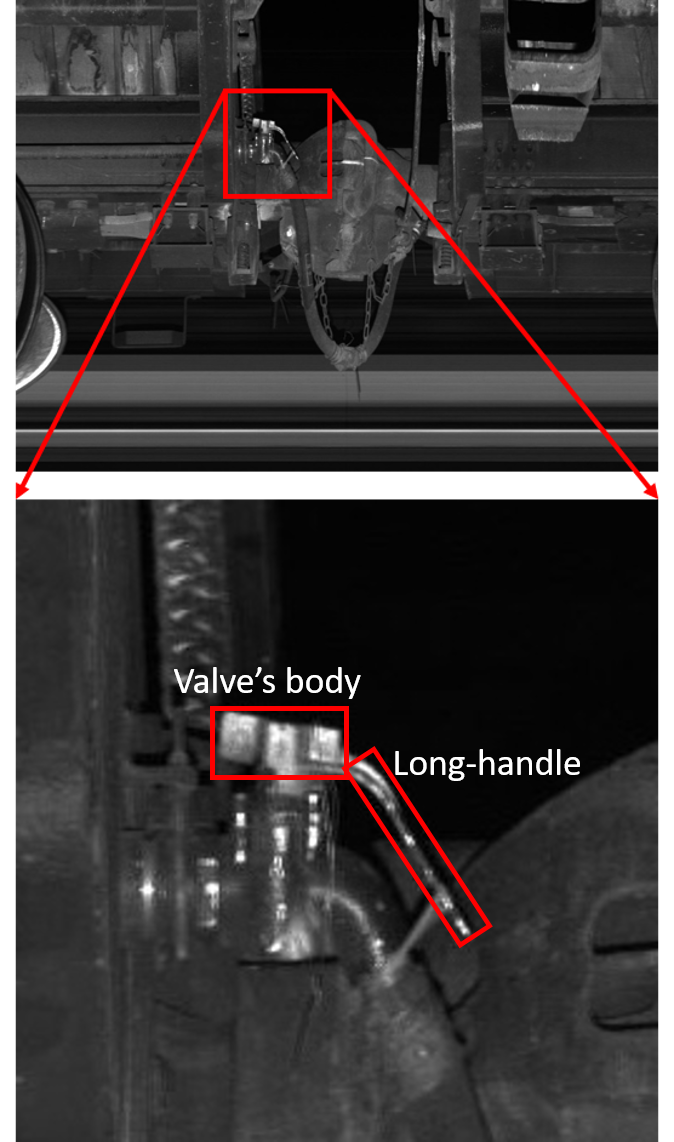} \caption{Normal Valve.}
\end{subfigure}  
\begin{subfigure}[b]{0.225\textwidth}
\includegraphics[width=\textwidth]{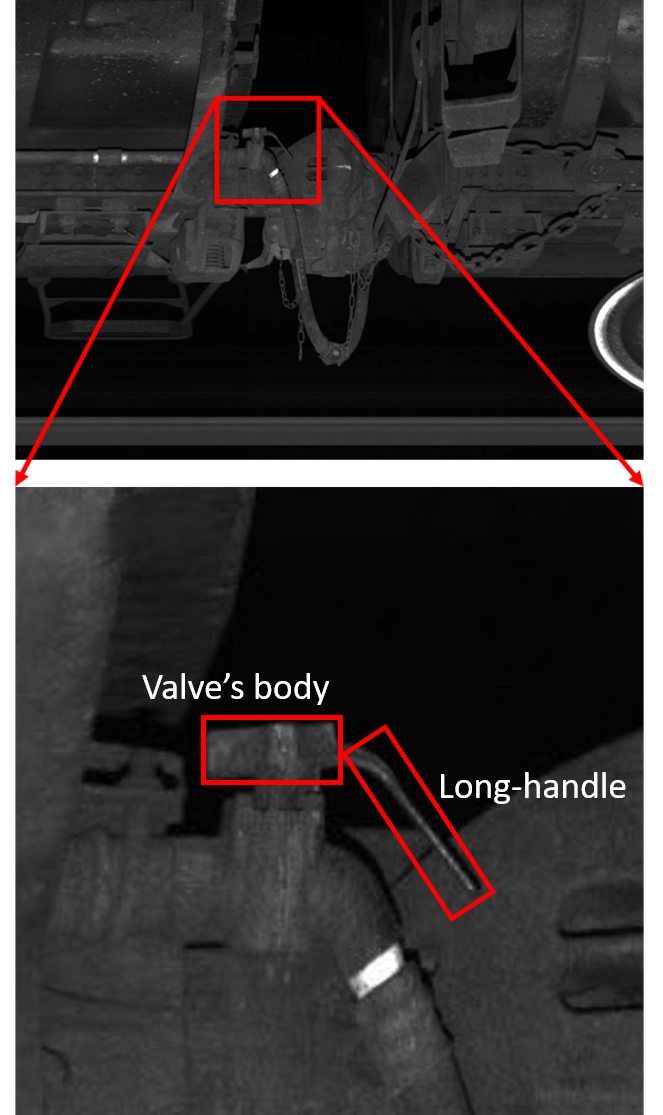}
\caption{Faulty Valve.}
\end{subfigure}  
\caption{We zoom in the ROI for visualization purposes. In a normal (non-faulty) valve, the valve's main body is facing downwards towards the valve long-handle as shown in (a). On the contrary, a valve is considered faulty if the valve's main body is turned upwards towards the valve long-handle or at the same height as shown in (b).}
\label{fig::valve_example_crop}
\end{figure}

\end{section}

\begin{section}{Our Approach}\label{sec::our_approach}

We utilize the images from the VTIS, with additional data augmentation, for image segmentation and fault classification purposes. Deep convolutional neural networks, such as U-Nets \cite{ronneberger2015u}, ResNets \cite{resnet_CVPR} and DenseNets \cite{densenet_CVPR}, are popular models when dealing with image segmentation and classification problems. We use these as our baseline models to compare and contrast our proposed hybrid approach - FaultNet - in terms of different metrics such as Dice coefficient and Intersection over Union (IoU) for segmentation and detection accuracy for fault detection.

The proposed technique, FaultNet, is a multi-phase  hybrid approach that integrates DL based high resolution image segmentation with computer vision based image processing to automatically detect faulty valves in rails. This is done by first performing an accurate high resolution segmentation of the valve using a two-step U-Net based segmentation. Thereafter, the segmented mask is used as an input to our image processing module that estimates the angle of the valve's main body with respect to the valve-handle. If this angle is detected to be more than $0\degree$, it is considered as a normal valve and vice-versa.

\begin{subsection}{U-Net based two-step segmentation}\label{subsec::faultNet}
For semantic segmentation, a two-step U-Net is used to extract the valve. U-Net is a CNN architecture for fast and precise image segmentation and is known to give great results for foreground-background segmentation applications. The architecture consists of a contracting encoder to analyze the entire image followed by a symmetric expanding decoder to produce an accurate segmentation of the object of interest. It also contains shortcuts, sometimes called skip connections, to preserve the pixelwise information at varying image resolutions. Due to limited memory and computation, the standard U-Net architecture usually downsamples the entire image to a smaller resolution and performs image segmentation on the downsampled data. Thereafter, the segmented masks are upsampled to the original resolution to produce final segmentation. The initial downsampling of the input image results in less accurate segmentation, especially at edges. In the present context, it is crucial to avoid these mistakes as a single pixel segmentation error may lead to an erroneous estimation of the valve-angle.  
\begin{figure}[t!]
\centering
\includegraphics[width=0.45\textwidth]{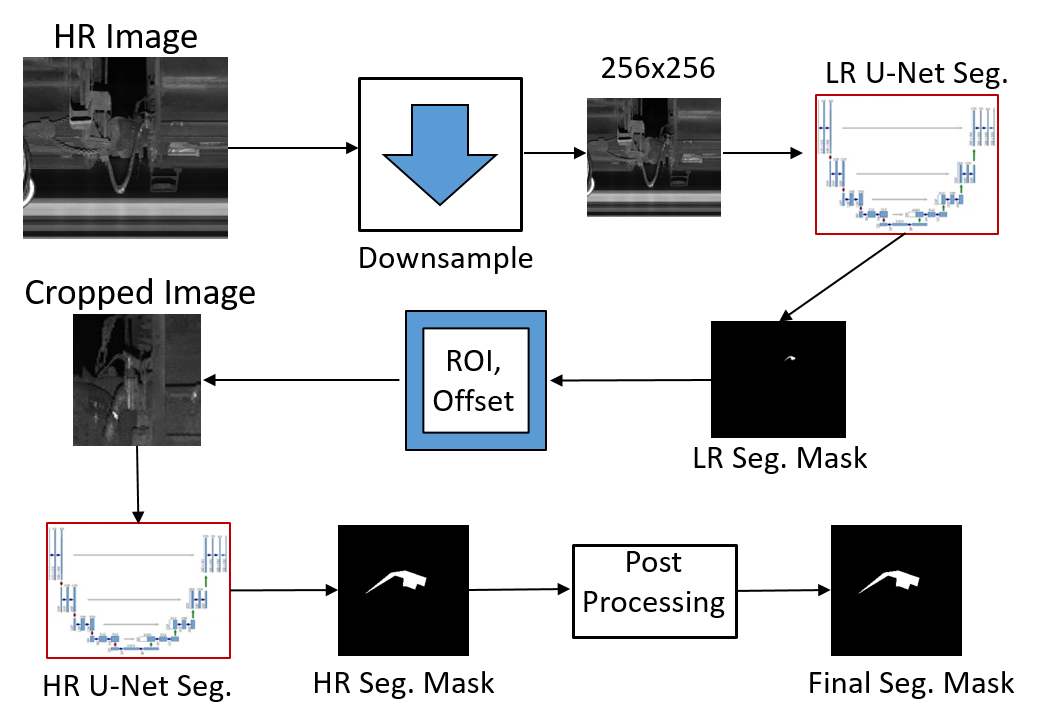}
\caption{Our Two-Step U-Net based inference process. The image is fed into the first network to find the ROI and then the cropped image is used by second image to segment out the full resolution object of interest.}
\label{fig::unet_ours_inference}
\end{figure}

We address this critical issue by exploiting our camera-scene setup. Since the cameras are stationary and the rails are at the same distance from the camera, the valves' size is approximately the same in each image. Thus, we do not need to account for same objects having varying sizes due to their distance from the camera locations. Leveraging on this fact, we perform a two-step U-Net based segmentation that first computes an ROI and then crops this region for performing segmentation on full (and cropped) resolution for a more accurate segmentation of the object of interest. 

In the standard U-Net approach, the dataset is first normalized and downsampled to the desired dimensions. Thereafter, the images and masks are used to train the standard U-Net. In our two-step approach, the dataset is also normalized and downsampled for training the first U-Net. The output mask is upsampled to original resolution and the ROI is used to create and save the cropped dataset for training the second U-Net. The offset information, per image, is also saved. This cropped dataset is also normalized before the training step. After the inference, the cropped segmentation of the second U-Net, along with the saved offset information is used to compute the full resolution segmented mask. A simple thresholding is applied on the segmented mask to obtain the final output binary mask.

In FaultNet, both the U-Nets use Adam algorithm with an initial learning rate of $1e-4$ and binary cross entropy as the loss function. The model is trained with images in mini-batches of $2$ and the best model is selected after $200$ epochs. The training and inference process for our image segmentation architecture is shown in Fig. \ref{fig::unet_ours_inference}. After performing the two-step segmentation step, image processing tools are used to crop the portion of the potential faulty valve in the images. The cropped images are saved and fed to the computer vision based module for fault detection.
\end{subsection}

\begin{subsection}{Fault detection}\label{subsec::fault_detection}
\begin{figure}[tpb]
\centering
\begin{subfigure}[b]{0.225\textwidth}
\includegraphics[width=\textwidth]{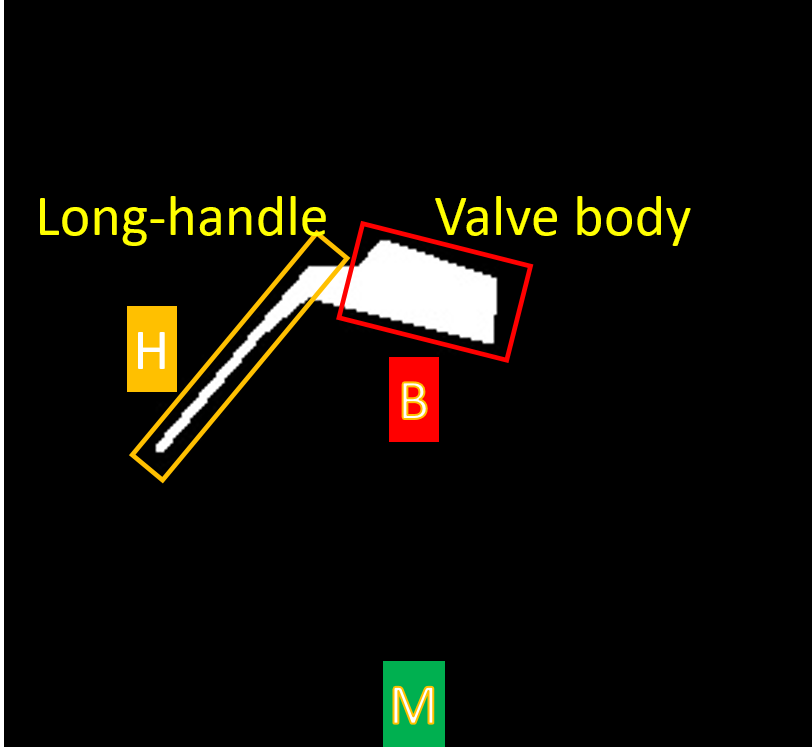} \caption{Two component detection.}
\end{subfigure}  
\begin{subfigure}[b]{0.225\textwidth}
\includegraphics[width=\textwidth]{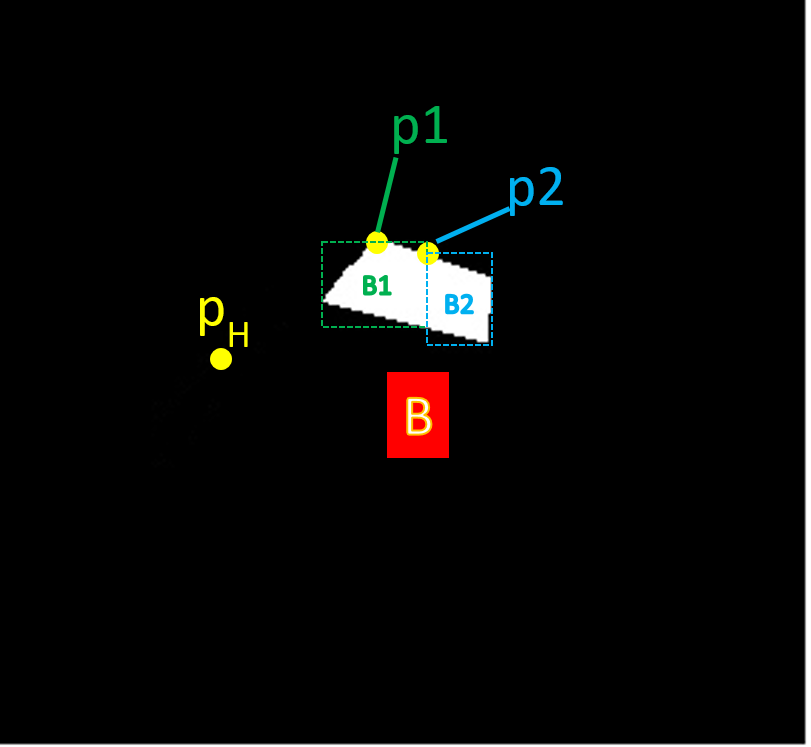} \caption{Faulty Valve.}
\end{subfigure}  
\begin{subfigure}[b]{0.225\textwidth}
\includegraphics[width=\textwidth]{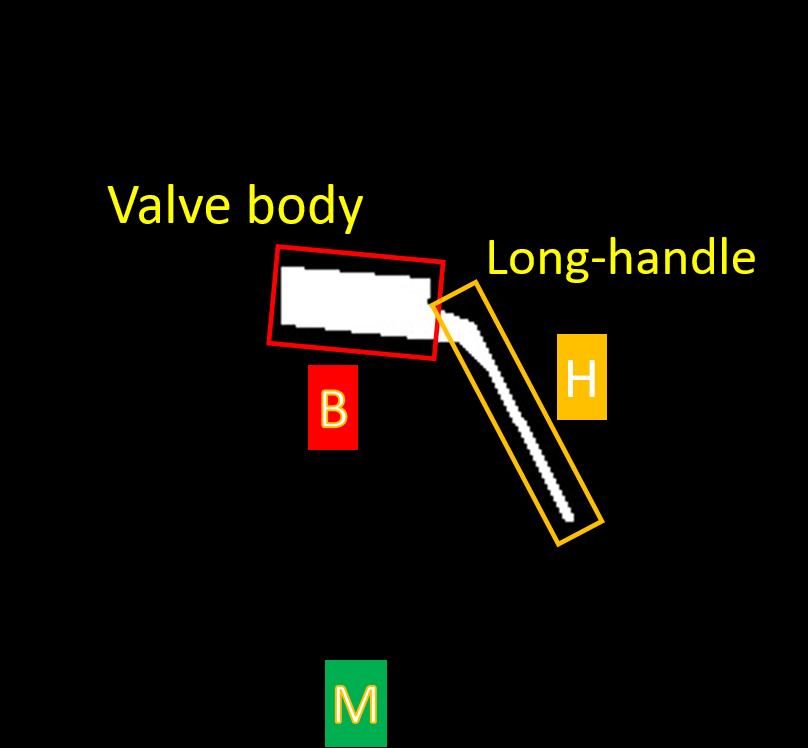} \caption{Two component detection.}
\end{subfigure}  
\begin{subfigure}[b]{0.225\textwidth}
\includegraphics[width=\textwidth]{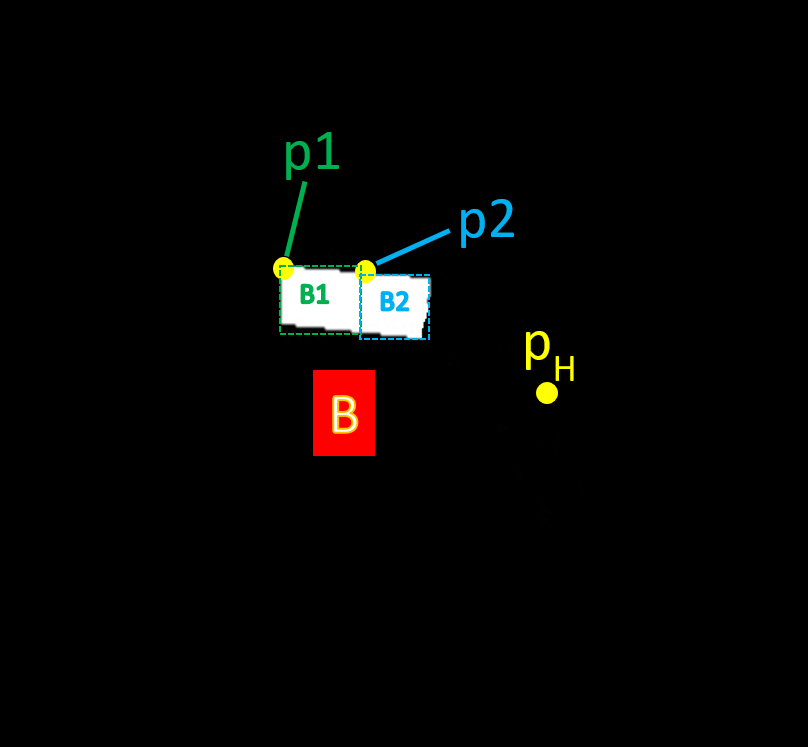}
\caption{Normal Valve.}
\end{subfigure} \caption{(a),(c) We perform two component detection on the binary mask, \textbf{M}, to detect the valve's main body and long-handle. We further use the valve's main body segmentation mask, \textbf{B}, to detect the top-most points, $p_1$ and $p_2$ lying on the two halves. (b) The valve is classified as faulty if the top-most point, $p_1$ lies in between $p_2$ and $p_H$. (d) On the contrary, if $p_2$ lies in between $p_1$ and $p_H$, and $p_1$ is strictly above $p_2$, the valve is classified as normal.}
\label{fig::img_processing_module}
\end{figure}

For the last phase in FaultNet, the cropped and segmented masks, \textbf{S}, are fed to an image processing module that detects whether the valve is faulty. The performance of the proposed technique is compared with state-of-the-art techniques for image classification such as ResNet \cite{resnet_CVPR} and DenseNet \cite{densenet_CVPR}. 

The proposed image processing module for fault detection is illustrated in Algorithm~\ref{algo::img_processing_module}. First, appropriate morphological functions - erode and dilate are used to remove any noisy segmentation that may have been output from the segmentation step. Thereafter, we retain the segmentation with most connected pixels using OpenCV based contour detection as a binary mask, \textbf{M}.

\begin{algorithm}[t]
\SetAlgoLined
 Input: segmented mask, \textbf{S}\;
 \textbf{M} $\gets$ ProcessMask(\textbf{S})\;
 \textbf{B}, \textbf{H} $\gets$ FindValveComponents(\textbf{M})\;
 $p_H$ $\gets$ FindLongHandleCentre(\textbf{H})\;
 $p_1$ $\gets$ FindTopMostPoint(\textbf{B})\;
 \textbf{B1}, \textbf{B2} $\gets$ DivideTwoHalves(\textbf{B})\;
 $p_2$ $\gets$ FindTopMostPoint(\textbf{B2})\;
 \uIf{($p_1[x] -p_H[x]$)($ p_1[x] - p_2[x]$) $\leq$ $0$}{
  	IsDefective = True \;
  	}
\uElseIf {$p_2[y] == p_1[y]$ }{
		IsDefective = True\;
}
\uElse{
IsDefective = False\;
}
 \KwResult{return IsDefective}
\caption{Our Computer Vision based Module to detect faulty valves.}
\label{algo::img_processing_module}
\end{algorithm}

The segmented mask, \textbf{M}, is first used for two-component detection - to detect the valve's main body and handle as shown in Fig.~\ref{fig::img_processing_module}(a,c). As the valve's long handle and main body occupy similar width, we first estimate a bounding box for the entire region and then divide the segmentation into two equally wide components. As the long handle's section contains mostly background region, we use the foreground to background ratio to identify which component contains the long-handle, \textbf{H} and valve's main body, \textbf{B}. 

The valve's main-body segmentation, \textbf{B}, is further processed to detect the topmost point, $p_1$, lying on the segmented mask. The mask is further divided into two equally wide boxes - \textbf{B1} and \textbf{B2}. \textbf{B1} contains this topmost point, $p_1$. The remaining segment, \textbf{B2} is processed to identify its top most point, $p_2$. If $p_1$ lies in between $p_H$ and $p_2$, as shown in Fig.~\ref{fig::img_processing_module}(b), it is classified as a faulty valve. On the contrary, if $p_2$ is detected to be in the middle, we perform further analysis to ensure the valve's body is indeed tilted towards the long-handle. If the points $p_1$ is strictly higher than $p_2$, the valve is classified as normal as shown in Fig.~\ref{fig::img_processing_module}(d).

This multi-phase segmentation and post-processing based technique gave us the best results as will be demonstrated in the following section.
\end{subsection}

\end{section}

\begin{section}{Experimental Results}\label{sec::results}
\begin{subsection}{Setup and Dataset Description}
In this study, we leverage on two different datasets - SMRT dataset and Rail-valve dataset. All the images used in our experiments are actual images that are collected by a COTS VTIS. The Singapore Mass Rapid Transit (SMRT) dataset images consist of rail engine consisting of items such as axle bolts, break, gasket, and screws. We use this dataset to demonstrate our two-step segmentation's superior performance over traditional U-Net segmentation technique. The SMRT dataset consists of $74$ images with each image having four different segmentation ROIs. The images have a resolution of $1024\times 1400$ pixels. An example image and the corresponding ROI components are shown in Fig.~\ref{fig::smrt_example}.

\begin{figure}
	\centering
	\includegraphics[width=.48\textwidth]{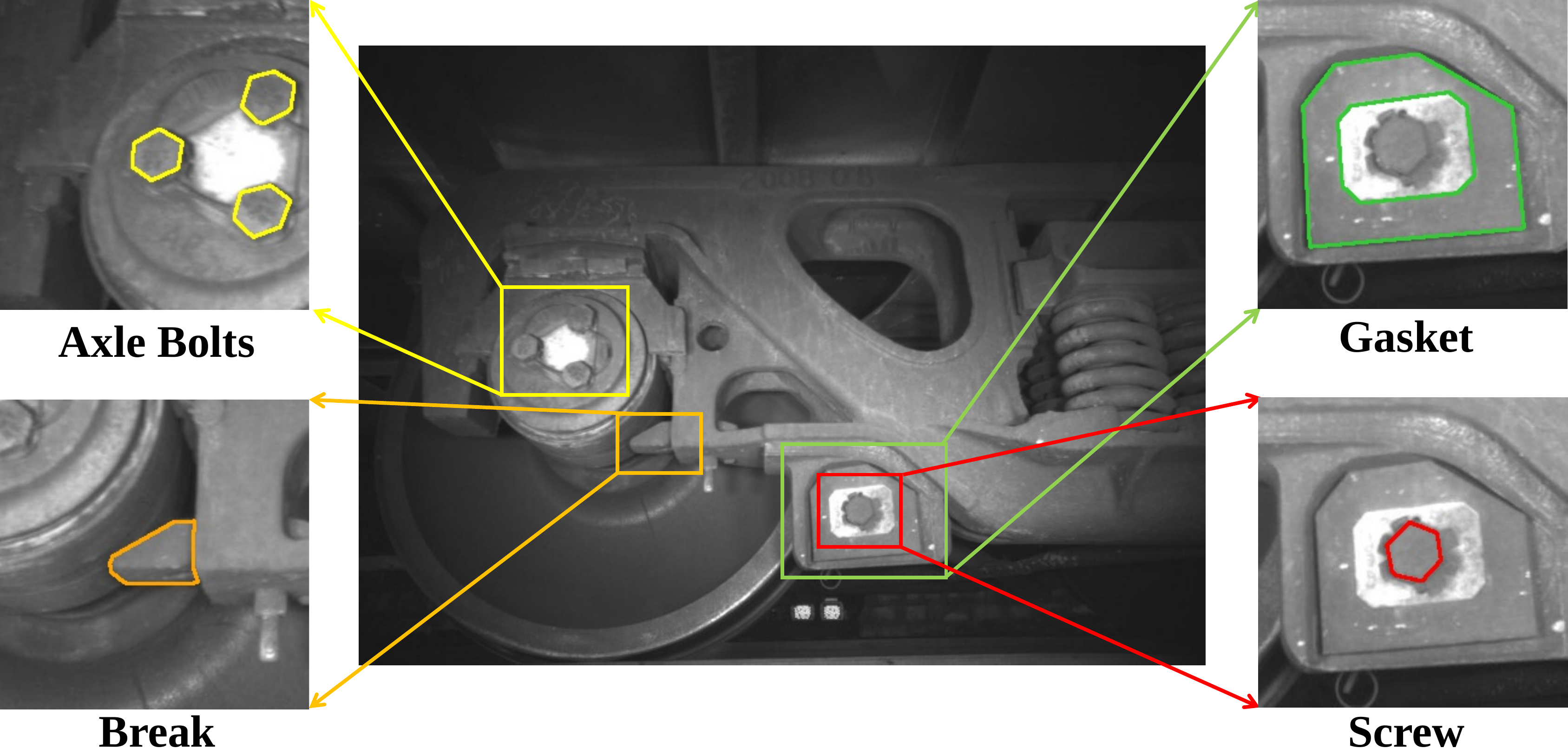}
	\caption{An example image from the SMRT dataset, with four segmentation ROI components.}
	\label{fig::smrt_example}
\end{figure}

The Rail-valve dataset consists of images of the normal and faulty valves. The Valve dataset consist of $73$ valve images. Among these, $31$ are normal valves and $42$ are faulty valves. The images also have $1024\times 1400$ pixel resolution. Fig.~\ref{fig::valve_example_crop} shows example images and the ROI components for the Rail-valve dataset.

Our first U-Net model normalizes and resizes the images to $256\times256$ to compute ROI and crops a $256\times256$ region around the valve ROI before performing the final high resolution segmentation. For the classification component in the second phase, the original and cropped images are resized to $224 \times 224$ before being fed to the ResNet/DenseNet for comparison with our approach. 

The experiments are run on an Intel Xeon Platinum $8170$ server with a Nvidia Tesla V100 GPU using Tensorflow $1.13.1$ on Ubuntu $16.04$. For the segmentation task, we use a $80$-$10$-$10$ split for training, validation, and testing. Meanwhile for the classification task, the dataset is divided into training and testing set with $75\%$ allocated for training and $25\%$ for testing for ResNet and DenseNet model evaluation. Further, $4$-fold cross validation is performed on the classification models. Since our Image processing module is purely computer vision based, all images are used for testing purposes.
\end{subsection}
\begin{table*}[t]
\centering
\caption{Dice Coefficient and IoU comparisons between classical U-Net and our segmentation approach on the SMRT dataset.}
\setlength{\extrarowheight}{3pt}
{ \begin{tabular}{ |c|c||c|c||c|c||c| } \hline 
{Segment} & {Architecture} &  Train Dice & Train IoU & Validation Dice &  Validation IoU & Test IoU \\ \hline 
\multirow{3}{*}{Axle bolts} & U-Net-$256$ & $0.8457$ & $0.7452$ & $0.8759$ & $0.7794$ & $0.7832$  \\  \cline{2-7}
& U-Net-$512$ & $0.8845$ & $0.8144$ & $0.9129$ & $0.8399$ & $0.8077$  \\  \cline{2-7}
& {Our Approach} & $\bf{0.9311}$ & $\bf{0.8721}$ & $\bf{0.9324}$ & $\bf{0.8733}$ & $\bf{0.8336}$  \\ \hline

\multirow{3}{*}{Break} & U-Net-$256$ & $0.8938$ & $0.8089$ & $0.9119$ & $0.8382$ & $0.7527$  \\  \cline{2-7}
& U-Net-$512$& $0.9369$ & $0.8816$ & $0.9423$ & $0.8910$ & $0.7575$  \\  \cline{2-7}
& {Our Approach} & $\bf{0.9676}$ & $\bf{0.9375}$ & $\bf{0.9615}$ & $\bf{0.9259}$ & $\bf{0.7991}$  \\ \hline

\multirow{3}{*}{Gasket} & U-Net-$256$ & $0.9270$ & $0.8642$ & $0.9192$ & $0.8507$ & $0.8946$  \\  \cline{2-7}
& U-Net-$512$ & $0.9547$ & $0.9138$ & $0.9438$ & $0.8938$ & $0.8905$  \\  \cline{2-7}
& {Our Approach} & $\bf{0.9573}$ & $\bf{0.9184}$ & $\bf{0.9571}$ & $\bf{0.9179}$ & $\bf{0.9066}$  \\ \hline

\multirow{3}{*}{Screws} & U-Net-$256$ & $0.8974$ & $0.8144$ & $0.8500$ & $0.7416$ & $0.8609$  \\  \cline{2-7}
& U-Net-$512$ & $0.9420$ & $0.8905$ & $0.8763$ & $0.7827$ & $0.8679$  \\ \cline{2-7}
& {Our Approach} & $\bf{0.9689}$ & $\bf{0.9399}$ & $\bf{0.8854}$ & $\bf{0.7979}$ & $\bf{0.8757}$  \\ \hline
\end{tabular}
}
\label{table::SMRT}
\end{table*}
\begin{subsection}{Image Segmentation}
\begin{figure}
	\centering
	\includegraphics[width=.48\textwidth]{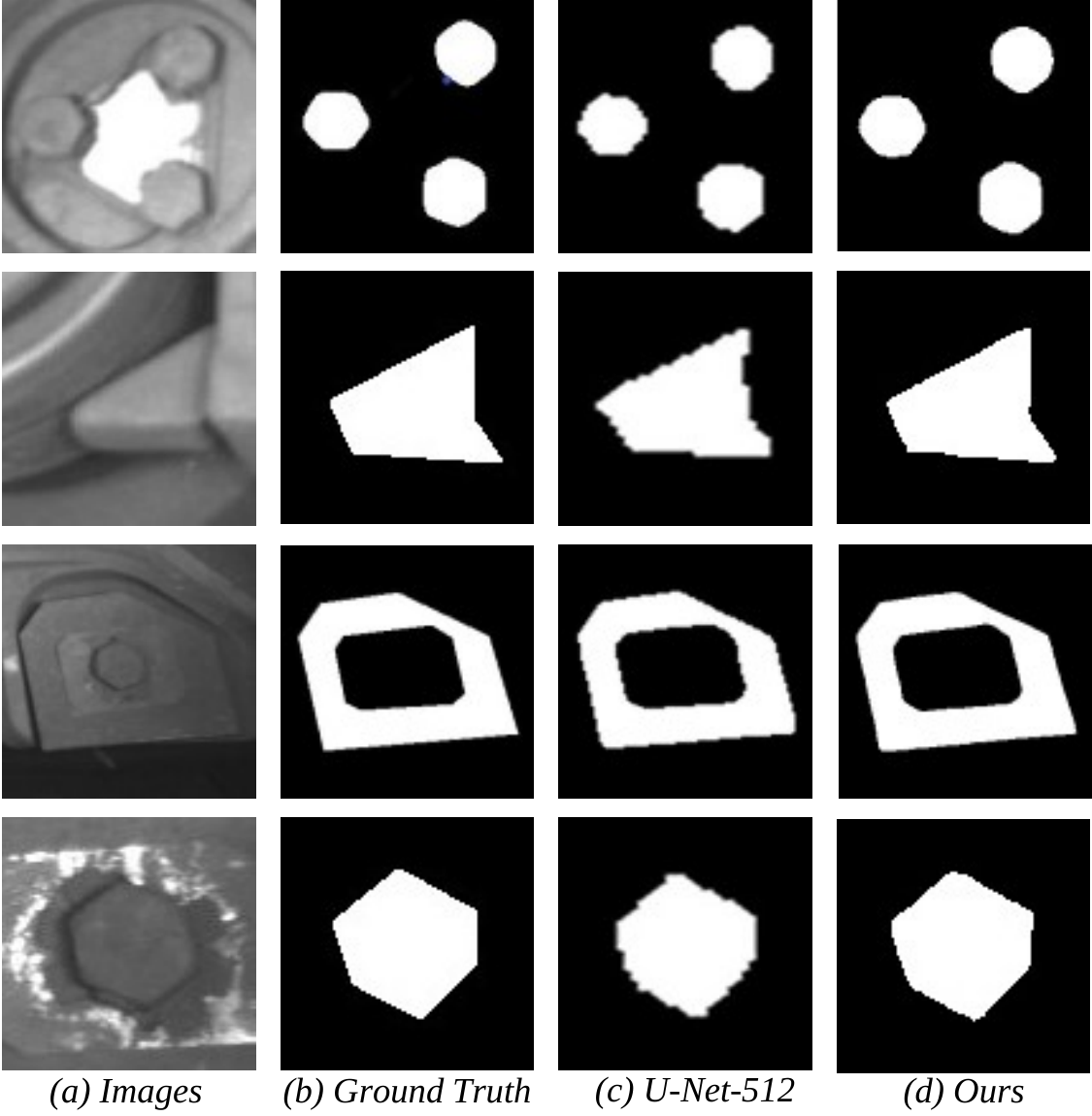}
	\caption{SMRT dataset segmentation on the four ROI components (axle bolts, break, gasket, and screws). (a) shows the images of the components. (b) shows the ground truth masks for the components. (c) shows the predicted masks by U-Net-512. (d) shows the predicted masks by our two-step segmentation approach. Our approach produces sharper, more consistent and accurate edges for object segmentation.}
\label{fig::smrt_pred_compare}
\end{figure}
A sample image-segment and their corresponding masks from SMRT dataset are shown in Fig.~\ref{fig::smrt_pred_compare}. We compare our segmentation results with original U-Net of $256\times256$ resolution (denoted as U-Net-256) and $512\times512$ resolution (denoted as U-Net-512) to demonstrate our technique's superior performance. We divide the dataset into $80\%-10\%-10\%$ training-validation-test split. In order to generate more images for the training, we also perform data augmentation techniques such as cropping, flipping, scaling and rotation on the original images. We compute the training and validation Dice Coefficient, and IoU and test IoU for comparison purposes as shown in Eq.~\ref{eq::metrics}. 
\begin{eqnarray} \label{eq::metrics}
Dice(X,Y)&=&\frac{2\times|X\cap Y|}{|X| + |Y|}; \nonumber \\
IoU(X,Y)&=&\frac{|X\cap Y|}{|X\cup Y|}
\end{eqnarray}
where X and Y represent the output binary mask and groundtruth binary mask respectively.
\end{subsection}

\begin{subsection}{Fault Detection}
For fault detection, we compare our results with state-of-the-art classification techniques - ResNet and DenseNet. For fairness, we perform the classification, for both techniques, on full images and the cropped ROI that only contain the valve. We expect the results to be better for cropped images dataset as it removes the redundant regions from the images. We normalize the images according to the mean and standard deviation of images in the ImageNet training dataset. We also perform data augmentation for both ResNet and DenseNet architectures. The images in the training dataset are augmented using standard parameters such as rotation up to $30\degree$, vertical and horizontal shift range of $0.2$, shear = $0.2$, zoom = $0.2$, and vertical and horizontal flips.

The weights of the ResNet and DenseNet model are initialized from a model trained on ImageNet \cite{ImageNet2015}. In this work, the final fully connected layer is replaced with a small customized CNN model that consists of two layers. The number of units in the two layers are $2048$ and $2$ for ResNet50 and $1024$ and $2$ for DenseNet-121, respectively. The final fully connected layer corresponds to the two-classes representing the normal and faulty valve cases. For each of the models, the training is performed for $200$ epochs and one with the lowest validation loss is selected as the best model. 

As explained in Sec.~\ref{subsec::fault_detection}, our hybrid approach uses the output segmented mask for fault detection. We process each image's mask as described in Algorithm~\ref{algo::img_processing_module} to detect faulty valves. 
\end{subsection}

\begin{table}[htbp]
\centering
\caption{Dice Coefficient and IoU comparisons on the Rail-Valve dataset.}
\setlength{\extrarowheight}{3pt}
{ \begin{tabular}{ |c|c|c|c|c|c| } \hline 
{Architecture} &  Train & Train & Val. &  Val. & Test \\ 
               &  Dice & IoU & Dice & IoU & IoU \\ \hline 
U-Net-$256$ & $0.7957$ & $0.6618$ & $0.7857$ & $0.6476$  & $0.6513$ \\ \hline
U-Net-$512$ & $0.8824$ & $0.7898$ & $0.8572$ & $0.7503$ & $0.6635$ \\ \hline
{Our Approach} & $\bf{0.9421}$ & $\bf{0.8907}$ & $\bf{0.8965}$ & $\bf{0.8126}$ & $\bf{0.8417}$  \\ \hline
\end{tabular}
}
\label{table::AC1_unet}
\end{table}
\begin{figure}[htpb]
\centering
\begin{subfigure}[b]{0.115\textwidth}
\includegraphics[width=\textwidth]{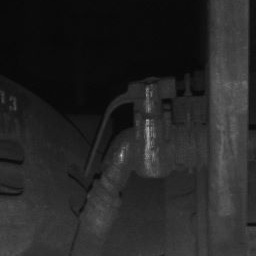} 
\end{subfigure}  
\begin{subfigure}[b]{0.115\textwidth}
\includegraphics[width=\textwidth]{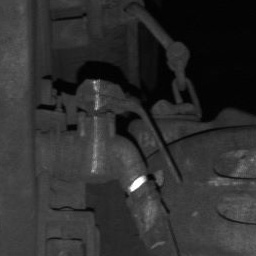} 
\end{subfigure}  
\begin{subfigure}[b]{0.115\textwidth}
\includegraphics[width=\textwidth]{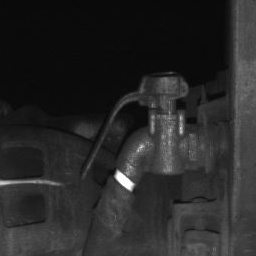} 
\end{subfigure}  
\begin{subfigure}[b]{0.115\textwidth}
\includegraphics[width=\textwidth]{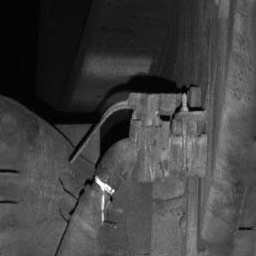}
\end{subfigure}
\begin{subfigure}[b]{0.115\textwidth}
\includegraphics[width=\textwidth]{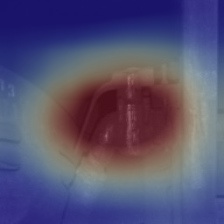} 
\end{subfigure}  
\begin{subfigure}[b]{0.115\textwidth}
\includegraphics[width=\textwidth]{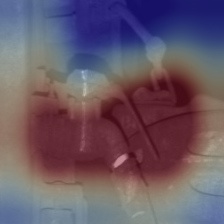} 
\end{subfigure}  
\begin{subfigure}[b]{0.115\textwidth}
\includegraphics[width=\textwidth]{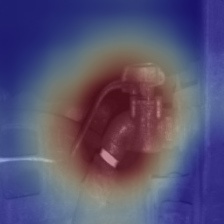} 
\end{subfigure}  
\begin{subfigure}[b]{0.115\textwidth}
\includegraphics[width=\textwidth]{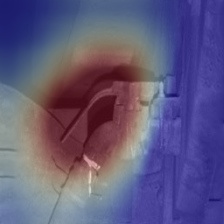}
\end{subfigure}\caption{We display a few examples where ResNet makes classification errors while our approach classifies them correctly. We also display the heatmaps for the classification. We can observe that sometimes the model does not focus on the valve's main-body and handle positioning for classification.}
\label{fig::fault_detection_ims1}
\end{figure}
\begin{subsection}{Results}
\textbf{Segmentation}:~ Table \ref{table::SMRT} shows the results for SMRT dataset. We use Dice coefficient and Intersection over Union (IoU) as our performance metrics as shown in Eq.~\ref{eq::metrics}. We display results for 4 ROI components - axle bolts, break, gasket, screws. Our approach outperforms both U-Net-256 and U-Net-512 models. As expected, U-Net-512 outperforms U-Net-256 in all training, validation, and testing phases. This is within expectations as U-Net-256 uses a lower resolution for performing segmentation. However, U-Net-512 also suffers from higher computational and memory resource overhead. In contrast, our two-step approach gives the best performance improving the dice coefficient and IoU significantly for training and validation dataset. The smaller the object, the more improvement we observe since we avoid losing information by cropping instead of downsampling the images in our approach. IoU on test data is also significantly improved. Moreover, our approach effectively reduces the computations by $50\%$ than the U-Net-512 model. A visual comparison on the segmentation results is also shown in Fig.~\ref{fig::smrt_pred_compare}. The two-step U-Net enables us to obtain sharp and consistent boundaries compared to the classical approach. We also demonstrate our superior performance on Rail-Valve dataset and the corresponding results are shown in Table~\ref{table::AC1_unet}.

\begin{figure}[tpb]
\centering
\begin{subfigure}[b]{0.115\textwidth}
\includegraphics[width=\textwidth]{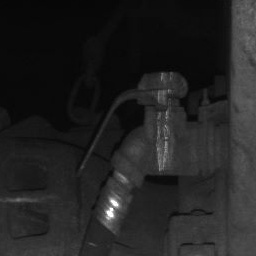}
\end{subfigure}  
\begin{subfigure}[b]{0.115\textwidth}
\includegraphics[width=\textwidth]{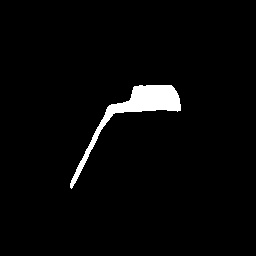}
\end{subfigure} 
\begin{subfigure}[b]{0.115\textwidth}
\includegraphics[width=\textwidth]{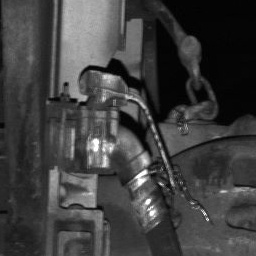}
\end{subfigure}  
\begin{subfigure}[b]{0.115\textwidth}
\includegraphics[width=\textwidth]{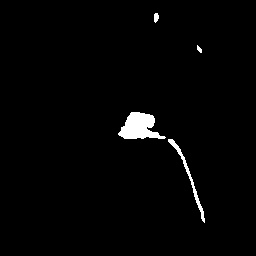}
\end{subfigure}\caption{We display the two cases where our approach makes the wrong classification. Due to a flat and noisy mask output from our first phase, the image processing module detects them to be faulty valves.}
\label{fig::fault_detection_ims2}
\end{figure}

\textbf{Fault Detection}:~ We perform fault detection using three methods - ResNet, DenseNet, and FaultNet. We use fault detection accuracy as our performance metric. The results are illustrated in Table~\ref{table::detection_res}. We use ResNets and DenseNets on both full image and cropped images for a comprehensive comparison. It can be observed that the performance of the FaultNet is significantly superior to both ResNet and DenseNet. Detecting accurate angles between the valve's main body and long-handle enables us to improve the fault detection performance significantly. ResNet and DensNet based approaches result in multiple skips and overkills. As stated before, a skip is extremely punishing as this will result in a faulty valve being identified as a normal step. This is unacceptable especially when it involves human lives. Our hybrid approach results in both zero skips and highest accuracy.
\begin{table}[htbp]
\centering
\caption{Faulty Valve detection Accuracy ($73$ images)}
\setlength{\extrarowheight}{3pt}
{ \begin{tabular}{ |c|c|c|c|c|c| } \hline 
{Approach} & Correct & Skips  & Overkills  &  Accuracy  \\ 
           & Detections &   &     &  \\ \hline 
ResNet50-full  & $65$ & $6$ & $2$ & $89.04\%$  \\ \hline
ResNet50-crop  & $66$ & $4$ & $3$ & $90.41\%$  \\ \hline
DenseNet121-full  & $67$ & $5$ & $1$ & $91.78\%$  \\ \hline
DenseNet121-crop  & $68$ & $3$ & $2$ & $93.15\%$  \\ \hline
{FaultNet}  & $\bf{71}$ & $\bf{0}$ & $\bf{2}$ & $\bf{97.26\%}$  \\ \hline
\end{tabular}
}
\label{table::detection_res}
\end{table}

Some escapes and overkills from the ResNet model along with their heatmaps are shown in Fig.~\ref{fig::fault_detection_ims1}. We observe that sometimes ResNet model does not focus on the correct areas - valve's main-body and long-handle to make the classification. We correctly identify these images as faulty (and normal) valves while the state-of-the-art techniques fail to do so. We also show the two errors we make in the SMRT dataset in Fig.~\ref{fig::fault_detection_ims2}. In the first valve-image, the valve upper body is detected to be flat, i.e. $p_1$ and $p_2$ are computed to be at same height. In the second image, light reflection off the valve saturates the region resulting in an extremely noisy output segmentation. We mark such images as faulty so they can be manually inspected rather than risking an escape.
\end{subsection}

\begin{subsection}{Limitations}
There are a few limitations of the proposed FaultNet technique. Firstly, this work focuses only on one type of classification namely faulty valves. Secondly, if valves are placed at an awkward angle such as facing away from the camera, our technique is unable to make a decision as the angle is difficult to infer from such images. Thirdly, we assume that an image contains a single valve and currently is not capable of handling multiple valves. However, this can be addressed by training the segmentation step with images of valves at different alignments and adding contour based constraints in first step segmentation to detect multiple valves, if present. 
\end{subsection}
\end{section}

\begin{section}{Conclusion}\label{sec::conclusion}
In this paper, a multi-phase deep learning technique is introduced for detecting faulty valves for automated rail inspection system. The first phase performs a full resolution semantic segmentation of the valve. In the second phase, a sophisticated image processing module is used to detect the valve angle with respect to its handle to perform fault classification. As shown, our approach enables us to focus on the ROI and results in significantly better performance than current state-of-the-art techniques. Moreover, our two-step segmentation approach is broadly applicable to various domains such as bio-medical, chip manufacturing and semiconductor analysis fields where the object of interest and the camera have a pre-defined distance between them enabling a higher resolution segmentation using our approach. In future, this approach can be extended to perform automated visual inspection for underground tunnels \cite{raman_icis, raman_iros_2019} and accurately segment objects in $3$D pointclouds \cite{raman_tcsvt_3D_prop, raman_apsipa_3D_prop, Pham_2019_CVPR}. We also intend to explore a more generalized approach that can automatically detect other types of defects, such as track faults, in the future.
\end{section}

\begin{section}{Acknowledgement}
This work was supported by Singapore-China NRF-NSFC Grant Ref No. NRF2016NRF-NSFC001-111.
\end{section}

\bibliographystyle{IEEEbib}
\bibliography{refs_DL}

\begin{thebibliography}{10}

\bibitem{mrt_website}
``{Average Daily Public Transport},''
  \url{https://data.gov.sg/dataset/public-transport-utilisation-average-public-transport-ridership},
\newblock Accessed: 2019-04-04.

\bibitem{wang2018deep}
J.~Wang, Y.~Ma, L.~Zhang, R.~X. Gao, and D.~Wu,
\newblock ``Deep learning for smart manufacturing: Methods and applications,''
\newblock {\em Journal of Manufacturing Systems}, vol. 48, pp. 144--156, 2018.

\bibitem{weimer2016design}
D.~Weimer, B.~Scholz-Reiter, and M.~Shpitalni,
\newblock ``Design of deep convolutional neural network architectures for
  automated feature extraction in industrial inspection,''
\newblock {\em CIRP Annals}, vol. 65, no. 1, pp. 417--420, 2016.

\bibitem{li2017intelligent}
Z.~Li, Y.~Wang, and K.~S. Wang,
\newblock ``Intelligent predictive maintenance for fault diagnosis and
  prognosis in machine centers: Industry 4.0 scenario,''
\newblock {\em Advances in Manufacturing}, vol. 5, pp. 377--387, 2017.

\bibitem{zenati2018efficient}
H.~Zenati, C.~S. Foo, B.~Lecouat, G.~Manek, and V.~R. Chandrasekhar,
\newblock ``Efficient gan-based anomaly detection,''
\newblock {\em arXiv preprint arXiv:1802.06222}, 2018.

\bibitem{Papaelias2008}
M.~Ph Papaelias, C.~Roberts, and C.~L. linDavis,
\newblock ``A review on non-destructive evaluation of rails: State-of-the-art
  and future development,''
\newblock {\em Proc. Institution Mech. Eng., Part F: J. Rail Rapid Transit},
  vol. 222, no. 4, pp. 367--384, 2008.

\bibitem{CLARK2004111}
R.~Clark,
\newblock ``Rail flaw detection: Overview and needs for future developments,''
\newblock {\em NDT \& E Intl.}, vol. 37, no. 2, pp. 111 -- 118, 2004.

\bibitem{Clark2002341}
R.~Clark, S.~Singh, and C.~Haist,
\newblock ``Ultrasonic characterisation of defects in rails,''
\newblock {\em Insight - Non-Destructive Testing and Condition Monitoring},
  vol. 44, no. 6, pp. 341--347, 2002.

\bibitem{EDWARDS2006468}
R.~S. Edwards, S.~Dixon, and X.~Jian,
\newblock ``Characterisation of defects in the railhead using ultrasonic
  surface waves,''
\newblock {\em NDT \& E Intl.}, vol. 39, no. 6, pp. 468--475, 2006.

\bibitem{Bentoumi2003}
M.~Bentoumi, P.~Aknin, and G.~Bloch,
\newblock ``On-line rail defect diagnosis with differential eddy current probes
  and specific detection processing,''
\newblock {\em Eur. Phys. J. Appl. Phys.}, vol. 23, no. 3, pp. 227--233, 2003.

\bibitem{Thomas2007}
H-M. Thomas, T.~Heckel, and G.~Hanspach,
\newblock ``Advantage of a combined ultrasonic and eddy current examination for
  railway inspection trains,''
\newblock {\em Insight - Non-Destructive Testing and Condition Monitoring},
  vol. 49, no. 6, pp. 341--344, 2007.

\bibitem{Marino2007}
F.~Marino, A.~Distante, P.~L. Mazzeo, and E.~Stella,
\newblock ``A real-time visual inspection system for railway maintenance:
  Automatic hexagonal-headed bolts detection,''
\newblock {\em IEEE Transactions on Systems, Man, and Cybernetics}, vol. 37,
  no. 3, pp. 418--428, May 2007.

\bibitem{Mandriota2004}
C.~Mandriota, M.~Nitti, N.~Ancona, E.~Stella, and A.~Distante,
\newblock ``Filter-based feature selection for rail defect detection,''
\newblock {\em Machine Vision and Applications}, vol. 15, no. 4, pp. 179--185,
  2004.

\bibitem{LinJie2009}
J.~Lin, S.~Luo, Q.~Li, H.~Zhang, and S.~Ren,
\newblock ``Real-time rail head surface defect detection: A geometrical
  approach,''
\newblock in {\em IEEE Intl. Symp. Indust. Electron.}, Jul. 2009, pp. 769--774.

\bibitem{shi2000normalized}
J.~Shi and J.~Malik,
\newblock ``Normalized cuts and image segmentation,''
\newblock {\em Departmental Papers (CIS)}, p. 107, 2000.

\bibitem{felzenszwalb2004efficient}
P.~F. Felzenszwalb and D.~P. Huttenlocher,
\newblock ``Efficient graph-based image segmentation,''
\newblock {\em International Journal of Computer Vision}, vol. 59, no. 2, pp.
  167--181, 2004.

\bibitem{haralick1985image}
R.~M. Haralick and L.~G. Shapiro,
\newblock ``Image segmentation techniques,''
\newblock {\em Computer vision, graphics, and image processing}, vol. 29, no.
  1, pp. 100--132, 1985.

\bibitem{arbelaez2011contour}
P.~Arbelaez, M.~Maire, C.~Fowlkes, and J.~Malik,
\newblock ``Contour detection and hierarchical image segmentation,''
\newblock {\em IEEE transactions on pattern analysis and machine intelligence},
  vol. 33, no. 5, pp. 898--916, 2011.

\bibitem{grady2006random}
L.~Grady,
\newblock ``Random walks for image segmentation,''
\newblock {\em IEEE Transactions on Pattern Analysis \& Machine Intelligence},
  , no. 11, pp. 1768--1783, 2006.

\bibitem{ronneberger2015u}
O.~Ronneberger, P.~Fischer, and T.~Brox,
\newblock ``{U-Net}: Convolutional networks for biomedical image
  segmentation,''
\newblock {\em Medical Image Computing and Computer-Assisted Intervention
  (MICCAI)}, vol. 9351, pp. 234--241, 2015.

\bibitem{badrinarayanan2015segnet}
V.~Badrinarayanan, A.~Kendall, and R.~Cipolla,
\newblock ``Segnet: A deep convolutional encoder-decoder architecture for image
  segmentation,''
\newblock {\em IEEE Transactions on Pattern Analysis and Machine Intelligence},
  2017.

\bibitem{maskrcnn}
K.~{He}, G.~{Gkioxari}, P.~{Dollár}, and R.~{Girshick},
\newblock ``Mask r-cnn,''
\newblock in {\em 2017 IEEE International Conference on Computer Vision
  (ICCV)}, Oct 2017, pp. 2980--2988.

\bibitem{Torsinopatent1}
J.~J. Cunningham, A.~E. Shaw, and M.~Trosino,
\newblock ``Automated track inspection vehicle and method,'' May 2000,
\newblock United States Patent 6064428.

\bibitem{Torsinopatent2}
M.~Trosino, J.~J. Cunningham, and A.~E. Shaw,
\newblock ``Automated track inspection vehicle and method,'' Mar. 2002,
\newblock United States Patent 6356299.

\bibitem{Ruvo20092333}
P.~De Ruvo, A.~Distante, E.~Stella, and F.~Marino,
\newblock ``A gpu-based vision system for real time detection of fastening
  elements in railway inspection,''
\newblock in {\em IEEE Intl. Conf. Image Processing (ICIP)}, Nov. 2009, pp.
  2333--2336.

\bibitem{Gilber2007}
X.~Gibert, A.~Berry, C.~Diaz, W.~Jordan, B.~Nejikovsky, and A.~Tajaddini,
\newblock ``A machine vision system for automated joint bar inspection from a
  moving rail vehicle,''
\newblock in {\em ASME/IEEE Joint Rail Conf. \& Internal Combustion Engine
  Spring Technical Conf.}, Mar. 2007, pp. 289--296.

\bibitem{YLi2014}
Y.~Li, H.~Trinh, N.~Haas, C.~Otto, and S.~Pankanti,
\newblock ``Rail component detection, optimization, and assessment for
  automatic rail track inspection,''
\newblock {\em IEEE Transactions on Intelligent Transportation Systems}, vol.
  15, no. 2, pp. 760--770, Apr. 2014.

\bibitem{ashish_tracknet}
A.~{James}, W.~{Jie}, Y.~{Xulei}, Y.~{Chenghao}, N.~B. {Ngan}, L.~{Yuxin},
  S.~{Yi}, V.~{Chandrasekhar}, and Z.~{Zeng},
\newblock ``Tracknet - a deep learning based fault detection for railway track
  inspection,''
\newblock in {\em 2018 International Conference on Intelligent Rail
  Transportation (ICIRT)}, Dec 2018.

\bibitem{Gibert2017}
X.~Gibert, V.~M. Patel, and R.~Chellappa,
\newblock ``Deep multitask learning for railway track inspection,''
\newblock {\em IEEE Transactions on Intelligent Transportation Systems}, vol.
  18, no. 1, pp. 153--164, Jan. 2017.

\bibitem{Giben2015}
X.~Giben, V.~M. Patel, and R.~Chellappa,
\newblock ``Material classification and semantic segmentation of railway track
  images with deep convolutional neural networks,''
\newblock in {\em IEEE Intl. Conf. Image Processing (ICIP)}, Sep. 2015, pp.
  621--625.

\bibitem{RoohiIJCNN}
S.~Faghih-Roohi, S.~Hajizadeh, A.~Núñez, R.~Babuska, and B.~De Schutter,
\newblock ``Deep convolutional neural networks for detection of rail surface
  defects,''
\newblock in {\em IEEE Intl. Joint Conf. Neural Networks (IJCNN)}, Jul. 2016,
  pp. 2584--2589.

\bibitem{resnet_CVPR}
K.~He, X.~Zhang, S.~Ren, and J.~Sun,
\newblock ``Deep residual learning for image recognition,''
\newblock in {\em The IEEE Conference on Computer Vision and Pattern
  Recognition (CVPR)}, June 2016.

\bibitem{densenet_CVPR}
G.~{Huang}, Z.~{Liu}, L.~V.~D. {Maaten}, and K.~Q. {Weinberger},
\newblock ``Densely connected convolutional networks,''
\newblock in {\em IEEE Conference on Computer Vision and Pattern Recognition
  (CVPR)}, July 2017, pp. 2261--2269.

\bibitem{ImageNet2015}
O.~Russakovsky, J.~Deng, H.~Su, J.~Krause, S.~Satheesh, S.~Ma, Z.~Huang,
  A.~Karpathy, A.~Khosla, M.~Bernstein, A.~C. Berg, and F.-F. Li,
\newblock ``Imagenet large scale visual recognition challenge,''
\newblock {\em Int. J. Comput. Vis.}, vol. 115, no. 3, pp. 115--211, 2015.

\bibitem{raman_icis}
R.~S. {Pahwa}, W.~{Kiat Leong}, S.~{Foong}, K.~{Leman}, and M.~N. {Do},
\newblock ``Feature-less stitching of cylindrical tunnel,''
\newblock in {\em IEEE/ACIS 17th International Conference on Computer and
  Information Science (ICIS)}, June 2018, pp. 502--507.

\bibitem{raman_iros_2019}
R.~S. Pahwa, K.~.Y. Chan, J.~Bai, V.~B. Saputra, M.~N. Do, and S.~Foong,
\newblock ``{Dense 3D Reconstruction for Visual Tunnel Inspection using
  Unmanned Aerial vehicle},''
\newblock in {\em IEEE/RSJ International Conference on Intelligent Robots and
  Systems (IROS)}, Nov 2019.

\bibitem{raman_tcsvt_3D_prop}
R.~S. Pahwa, J.~Lu, N.~Jiang, T.~T. Ng, and M.~N. Do,
\newblock ``{Locating 3D Object Proposals: A Depth-Based Online Approach},''
\newblock {\em IEEE Transactions on Circuits and Systems for Video Technology
  (TCSVT)}, vol. 28, no. 3, pp. 626--639, March 2016.

\bibitem{raman_apsipa_3D_prop}
R.~S. Pahwa, T.~T. Ng, and M.~N. Do,
\newblock ``{Tracking objects using 3D object proposals},''
\newblock in {\em Asia-Pacific Signal and Information Processing Association
  Annual Summit and Conference (APSIPA ASC)}, Dec 2017, pp. 1657--1660.

\bibitem{Pham_2019_CVPR}
Q-H. Pham, T.~Nguyen, B-S. Hua, G.~Roig, and S-K. Yeung,
\newblock ``{JSIS3D}: Joint semantic-instance segmentation of 3d point clouds
  with multi-task pointwise networks and multi-value conditional random
  fields,''
\newblock in {\em IEEE Conference on Computer Vision and Pattern Recognition
  (CVPR)}, June 2019.

\end{thebibliography}

\end{document}